\documentclass[12pt]{iopart}
\usepackage{url}            % simple URL typesetting
\usepackage{graphicx}
\usepackage{harvard}

\begin{document}
	
\title{3D segmentation of mandible from multisectional CT scans by convolutional neural networks}

\author{Bingjiang Qiu$ ^{1,2} $, Jiapan Guo$ ^{4, 6} $, J. Kraeima$ ^{1,3} $, R.J.H. Borra$ ^{2,5} $, M.J.H. Witjes$ ^{1, 3} $ and P.M.A. van Ooijen$ ^{1,2} $}

\address{
$^1 $3D Lab, 
$ ^2 $ Department of Radiology, 
$ ^3 $Department of Oral and Maxillofacial Surgery, 
$ ^4 $Center for Medical Imaging, 
$ ^5 $Department of Nuclear Medicine and Molecular Imaging, 
$ ^6 $Corresponding author \\
University Medical Centre Groningen, Hanzeplein 1, 9713GZ, Groningen, The Netherlands
}
\ead{j.guo@umcg.nl}
\vspace{10pt}
\begin{indented}
\item[]September 2018
\end{indented}
\begin{abstract} 
Segmentation of mandibles in CT scans during virtual surgical planning is crucial for 3D surgical planning in order to obtain a detailed surface representation of the patients bone. Automatic segmentation of mandibles in CT scans is a challenging task due to large variation in their shape and size between individuals. In order to address this challenge we propose a convolutional neural network approach for mandible segmentation in CT scans by considering the continuum of anatomical structures through different planes. The proposed convolutional neural network adopts the architecture of the U-Net and then combines the resulting 2D segmentations from three different planes into a 3D segmentation. We implement such a segmentation approach on 11 neck CT scans and then evaluate the performance. We achieve an average dice coefficient of $ 0.89 $ on two testing mandible segmentation. Experimental results show that our proposed approach for mandible segmentation in CT scans exhibits high accuracy. 
%, but also be efficiently accomplished by benefiting from  computing.
%Experimental results show that our proposed approach for mandible segmentation in CT scans is effective and superior to the existing approaches for mandible segmentation. 
%It is known that automated segmentation of medical images usually encounters challenges because of the large shape and size variations of anatomy between patients. To address this problem, we propose a novel neural network method for CT segmentation via considering anatomical structural continuum and different features in three directions (Axial, Coronal, Sagittal). The netword is inspired by that mandible has continuous structure and is a combination of three subnetworks. 
%%The proposed medical images segmentation is solved using. 
%In addition, our approach also achieves good performance for image segmentation under feeding all the skulls as inputs. Experimental results show that our method is effective and superior to the traditional methods. 
	
\end{abstract}

\section{Introduction}

\subsection{Motivation}

Three-dimensional virtual surgical planning (3D VSP) provides a precise and predictable method for bone related craniofacial tumor resection and free flap reconstruction of the reseted mandible. It is performed pre-operatively to determine the resection margins and cutting planes. The planning is translated towards the actual surgical procedure by the use of patient specific 3D printed guides. Computed tomography (CT) is the most commonly used modality for implementing such a process. Manual segmentation of the mandible in CT,
%regarding to a large number of dataset, 
however, leads to a tedious procedure in the clinical practice. Moreover, the structural complexity of mandibles and the considerable inter- and intra-rater variability make the segmentation of mandibles in CT scans challenging. 

%At present, three dimensional (3D) virtual planning of craniofacial tumor resection and free flap reconstruction is based on computed tomography (CT). Cutting and drilling guides are used in order to translate the 3D virtual plan towards the actual surgical procedure. It is essential to pre-operatively perform a 3D virtual planning to decide the resection margins and cutting planes in the mandible. 
%%There is one essential decision that should already be made pre-operatively during 3D virtual planning, which is the resection margins and then cutting planes in the bone of the mandible. 
%Thus, it is fundamental to segment the mandible from CT scans. 
%%Mandible bone segmentation from CT images is a fundamental and important work in 3D virtually plan. 
%However, manual segmentation of mandible is a time-consuming regarding to a large number of dataset, which leads to a tedious process in the clinical practice. It is also a challenge to segment mandible bone from their CT images because of structural complexity of mandible. Furthermore, manual segmentation is subject to considerable inter- and intra-rater variability. 

\subsection{Related works}

In the past decades, several semi-automatic and automatic methods have been developed to segment the mandible in CT scans.
%They can be categorized into two types, namely traditional and deep learning based approaches. several conventional methods, including automatic and semi-automatic methods, 
In multiple publications conventional methods have been proposed \cite{gollmer2012fully}\cite{torosdagli2017robust}\cite{chuang2017novel}\cite{abdi2015automatic}. A statistical shape model for segmentation of the mandible was presented by Gollmer el \cite{gollmer2012fully} in 2012. Torosdagli el \cite{torosdagli2017robust} proposed an image segmentation algorithm using 3D gradient-based fuzzy connectedness to segment mandibles. A registration-based semiautomatic mandible segmentation technique was proposed by Chuang el \cite{chuang2017novel} in 2017. The studies of Abdi el \cite{abdi2015automatic} show an automatic segmentation of mandible via collecting superior, inferior and exterior border in panoramic x-rays. 

Since 2013, convolutional neural networks (CNNs) have been successfully applied in computer vision tasks, for examples, image classification, super-resolution, and semantic segmentation. Medical image segmentation, as one of the research focuses in semantic segmentation, has developed exponentially due to the rapid evolution of CNNs. Recent advances in semantic segmentation \cite{long2015fully}\cite{badrinarayanan2017segnet}\cite{ronneberger2015Unet}\cite{yu2015multi}\cite{chen2018deeplab}\cite{lin2017refinenet}\cite{zhao2017pyramid}\cite{peng2017large}\cite{chen2017rethinking}\cite{garcia2017review} have enabled their applications to medical image segmentation. Fully convolutional network (FCN) is one of the first methods that introduced CNN into semantic segmentation and showed its ability on training an end-to-end approach for image segmentation with arbitrary input image sizes \cite{long2015fully}\cite{garcia2017review}. SegNet \cite{badrinarayanan2017segnet} and U-Net \cite{ronneberger2015Unet} are modified architectures based on FCN, that include an encoder and a decoder network. Both 3D-Unet \cite{cciccek20163d} and V-net \cite{milletari2016Vnet} are creative expansion techniques by taking into account the spatial information of medical images. 
The work in \cite{yu2015multi} proposed a multi-scale context aggregation module by dilated convolutions, which provides a solution to dense prediction problems in semantic segmentation. 
%Dilated convolutions \cite{yu2015multi} is a multi-scale context aggregation module. 
%The module makes the receptive fields larger without increasing the number of parameters and then obtains dense prediction results. 
The use of dilated convolutions enables the enlargement of receptive fields without the increment of parameters and the loss of resolution.
Deeplab \cite{chen2018deeplab} is an improved CNN architecture on semantic segmentation that uses dilated convolutions as well as conditional random fields. Refinenet \cite{lin2017refinenet} has a different architecture from that of the encoder-decoder mode. It makes use of all the information available along the down-sampling process to enable high-resolution prediction using long-range residual connections. 
The decoder part in Refinenet directly takes the feature maps of the encoder parts for fusion. PSPNet \cite{zhao2017pyramid} proposed a method of combining local and global information for dense prediction. An encoder-decoder structure with a large-scale convolution kernel was proposed by Peng \cite{peng2017large}. 
Most of the above mentioned CNN architectures have been proven to be effective on semantic segmentation of natural scene images. Such works \cite{long2015fully}\cite{badrinarayanan2017segnet}\cite{ronneberger2015Unet}\cite{yu2015multi}\cite{chen2018deeplab}\cite{lin2017refinenet}\cite{zhao2017pyramid}\cite{peng2017large}\cite{chen2017rethinking}\cite{garcia2017review} motivate our implementation of CNNs for automatic mandible segmentation in CT scans.
%Most importantly, these CNN methods have proven to be highly robust to varying image appearance, which motivates us to apply them to fully automatic segmentation of mandible in CT scans. 

Considering the applications of these CNN approaches in medical image segmentation for anatomical structures, we could feed the medical imaging data into the CNN architectures in two ways: direct 3D data and 2D slice-by-slice data. Medical imaging datasets usually have a large number of images at high resolution. On the one hand, during a 3D segmentation, small volumetric patches are produced to feed into the CNNs in order to deal with memory issues. In such a way, it, however, loses the anatomical structural information comparing to the full volumetric data. On the other hand, 2D slice-by-slice segmentation cannot take into consideration the spatial information in depth. Thus, we propose a CNN based approach to deal with such situations in medical image segmentations, particularly in the segmentation of mandible from CT neck scans. 
%In general, these previous methods of CNNs can be classified into two classes: direct 3D segmentation and 2D slice-by-slice segmentation. And medical images are usually high-resolution and full high-definition images. Commonly, medical data should be cut to smaller patches for training in direct 3D segmentation because the question of out of memory. However, it will miss some totally anatomical important structural information after cutting. And 2D slice-by-slice segmentation is also skipping spatial information in depth. For dealing with this drawbacks, we propose a architecture which is considering mandible anatomic structure and the computing capabilities. 
%The architecture is inspired by video objects segmentation. 

\subsection{Contribution}

In this work, we propose a CNN-based approach for the automatic 3D segmentation of mandibles in CT neck scans. Such an approach uses multisectional CT data from different orthogonal planes as input and then combines the 2D segmentation results from each plane in order to achieve a mandible segmentation in a three dimensional view. Such a design of the system enables the consideration of similarity and structural continuity of the mandible from different planes. Experimental results on 11 CT data demonstrate the effectiveness of the proposed approach.
% We train three CNNs to build a fully-automated system for the accurate segmentation of mandible. Unlike previous image segmentation networks, we use multisectional CT images as input data for considering the similarity and structural continuity of upper and lower slice images of mandible in the three planes and combine the feature information in the three directions.
%{\color{red}{}} 

To summarize, the main contributions of this work are as follows: 
%Experimental results suggest not only our best performing network outperforms a conventional segmentation method with hand-crafted features with a considerable margin, but also its performance does not significantly differ from an independent test images. To summarize, the main contributions of the paper are as follows: 
\begin{enumerate}
\item To the best of our knowledge, this is the first work that uses deep learning approach to effectively segment the mandible in CT scans. 
\item The proposed approach extracts discriminate features from three orthogonal planes (axial, coronal and sagittal) which take into account the structural continuity of mandibles from different views. 
\item The achieved experimental results demonstrate that the feasibility and effectiveness of the proposed approach in 3D mandible segmentation.
%, ever other segmentation complicated organ with continuous structure. 
\end{enumerate} 

%The remainder of  this paper is organized as follows. In Section \ref{section:methods}, we introduce our proposed approach for 3D mandible segmentation. Section \ref{section:experiments} describes implementation and experimental results of the proposed approach on 11 CT volumetric scans. Section \ref{section:discussion} presents a discussion and Section \ref{section:conclusion} draws conclusions. 

\section{Methods} \label{section:methods}

%\begin{figure}[!ht]
%\centering
%\subfigure[time] {\label{fig:time}\includegraphics[width=0.40\textwidth]{11.eps}}
%\caption{}
%\label{fig:time}
%\end{figure}

\subsection{Network architecture}

\begin{figure}
	\centering
	\includegraphics[width=380pt]{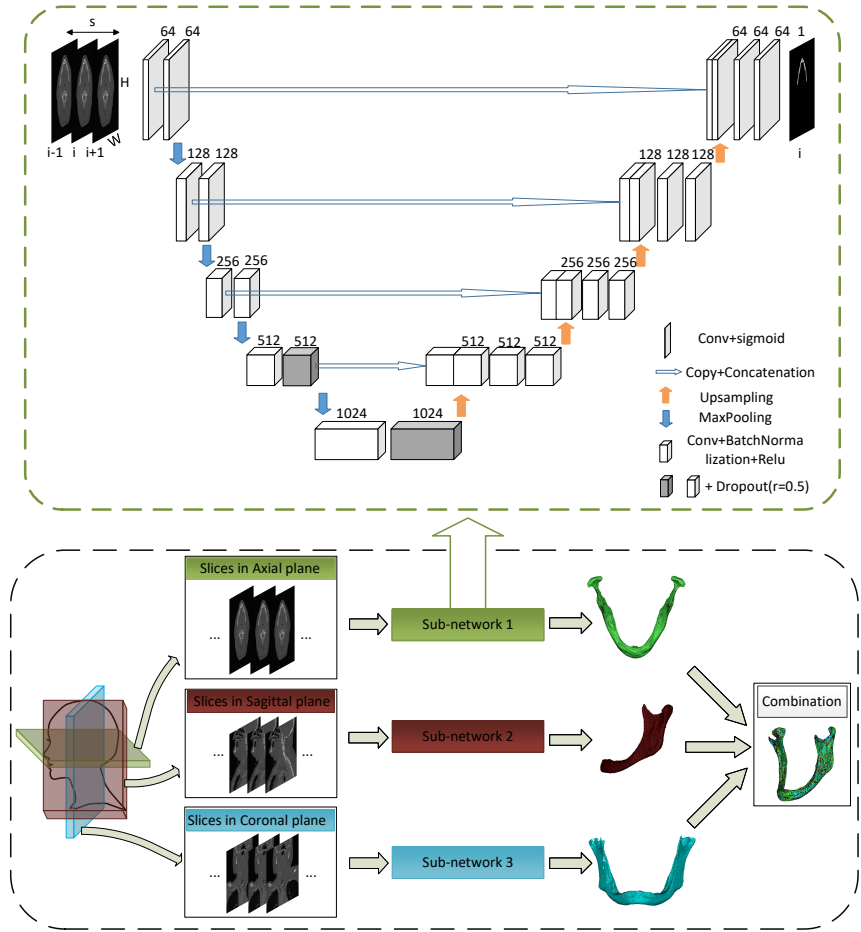}
	\caption{End-to-end architecture of the proposed convolutional neural network for 3D segmentation of the mandible, where $ H $, $ W $ and $ s $ represent the height, width and number of slices in the input layer, respectively. All the convolutional kernels have a size of $3\times3$.}
	\label{fig:architecture}
\end{figure}

In this work, we present a novel framework for the 3D segmentation of the mandible in CT scans. Figure \ref{fig:architecture} shows the architecture of the presented framework. Such a framework consists of three parallel channels which segment the mandible from different orthogonal directions. Each of the channels has the same architecture of the U-Net. The 2D segmentation results from each channel are then combined into a 3D segmentation. In the following, we elaborate on how we built such a framework. 

\subsection{Single-channel CNN model} 

In this section, we elaborate on the design of the single channel module. Figure 1 (top row) shows the architecture of the single-channel module, which is inspired by the architecture of the U-Net \cite{ronneberger2015Unet}. 
%We need to say about the input which is a mini 3D data with consecutive slices. 
%The single-direction CNN model we applied to our dataset is inspired by U-net \cite{ronneberger2015Unet}. 
In order to consider the similarity and continuity of upper and lower slice images of mandible, we apply multi-sectional images for training, which can maximally retain the structural information of the mandible. 
%The architecture shown in the top box of Figure \ref{fig:architecture} demonstrates the single direction architecture deep CNN is on the basis of U-net, which feeds multi-sectional images as inputs instead of single-sectional image. 
This single-channel network consists of encoding and decoding procedures of in total 23 convolutional layers, in which size of each convolutional kernel is $ 3 \times 3 $. The number of feature maps are listed on the top of each block that represents the convolutional layer. During the encoder procedure, max pooling layers are used to further enlarge the receptive fields. In order to achieve a pixel-wised classification, during the decoder procedure, upsampling layers are used and the feature maps are then concatenated with those of the same resolution from the decoder procedure. Different from that of the U-Net which concatenate the second feature maps from the same resolution in the encoder with those in the decoder, we pass the first block of feature maps to the decoder. Such a consideration helps to save computation resources. 
%This network consists of 23 convolutional layers that number of features in each layer is shown in Figure \ref{fig:architecture}. For encoder procedure, max pooling is applied to further enlarge the receptive field. Due to fully and adequately application of max pooling, the receptive field of the entire single-direction architecture can be very large, which guarantees good segmentation performance. 
%To improve the lack of information on the upsampling layers, concatenation of the same resolution is employed for learning the underlying features. But there is a bit change for saving computing resources, that we cascade the first block of convolution layer on the last three upsampling modules, as shown in the Figure \ref{fig:architecture}. 
%For decoder processing, concatenation of the same resolution 

Each of the convolutional layers is composed of a linear convolution, a batch normalization \cite{ioffe2015batch} and an element-wise nonlinear Relu function \cite{nair2010rectified}. Finally the resulting responses are turned into probability values using a sigmoid classifier \cite{lin2003study}. The output has the same size as that of the input image in the middle slice, as shown in Figure \ref{fig:architecture}. In general, the convolutional layers (illustrated by the cubic blocks in Figure \ref{fig:architecture}) can be expressed in the following equation: 
\begin{equation}
\label{eq:convolution}
L_i = \Gamma (L_{i-1} \otimes K_{ij} + B_{i}), 
\end{equation}
where $ L_{i} $ is an output of the $ i $-th layer and the number of feature maps for $ L_{i} $ is shown in the top of each layer in Figure \ref{fig:architecture}, $ \Gamma $ represents a nonlinear operator of Relu function, $ \otimes $ is a convolutional operator, $ K_{ij} $ means the $ j $-st feature kernel in the $ i $-st layer, and $ B_{i} $ denotes a bias of the $ i $-st layer. 
%We do not use pooling since it results in a shift-invariance property [86], which is not desired in segmentation tasks. Then we apply three layers of fully connected neurons of size 300, 200 and 2. 
%, which is a basis for structural continuity sensitive architectures,

%\textbf{In order to investigate thoroughly more context and structural information for image segmentation}, we use a loss function based on the Dice coefficient \cite{ghafoorian2017location}.
We use a loss function based on Dice coefficient which is commonly used to evaluate the performance of image segmentation tasks \cite{ghafoorian2017location}. Detailed information about Dice can be found in Section \ref{section:loss}. 

\subsection{Combination of the multi-sectional 2D mandible segmentation }

%To get the image features as much as possible, 
%With the images from only one direction, we lose shape information which can be observed from a different direction. 
The use of consecutive slices of scans helps improve the 2D segmentation of the mandible. However, on its own it does not provide satisfactory results since some parts of the mandible shapes are better observed via a different plane. Thus, we create a three channel network, of which each channel is fed with the CT data from a different orthogonal direction (axial, sagittal, coronal). We train these three channels separately using the corresponding data. 

\subsection{Loss metric}\label{section:loss}

%The Dice score is evaluated as: \cite{ghafoorian2017location}
The Dice similarity index, also called the Dice score, is often used to measure consistency between two samples\cite{ghafoorian2017location}. Therefore, it is widely applied as a metric to evaluate the performance of image segmentation algorithms. It is defined as: 
\begin{equation}
\label{eq:Dice}
{\rm Dice} = \frac{2|Y_{r} \cap Y_{p}|}{|Y_{r}|+|Y_{p}|}, 
\end{equation}
where $ Y_{r} $ is the reference standard, and $ Y_{p} $ is the predicted label from the network. Such a score has a value between $0$ and $1$, in which $ 0 $ means complete disagreement between the reference standard and the evaluated segmentation and $ 1 $ presents complete agreement. A differentiable technique has been used by Milletari \cite{milletari2016Vnet}, which we utilize for minimizing the loss (${\rm loss} = 1-{\rm Dice} $) between the two binary labels in training the presented model. 

\section{Experiments}\label{section:experiments}

\subsection{Data preparation}
%From each voxel neighborhood, we extract patches with three different sizes: $32 \times 32$, $64 \times 64$ and 128$\times$128. To reduce the computational costs, we down sample the larger two scales to 32$\times$32. Resulting patches for this procedure are demonstrated in Figure 2, for a negative and a positive sample, obtained from a FLAIR image. We included these three patches for both the T1 and FLAIR modalities for each sample. This results in a set of patches in three scales s1, s2 and s3, each consisting of two patches from T1 and FLAIR, as depicted in Figure 3.
%
%\subsection{Data}

\begin{figure}
	\centering
	\includegraphics[width=220pt]{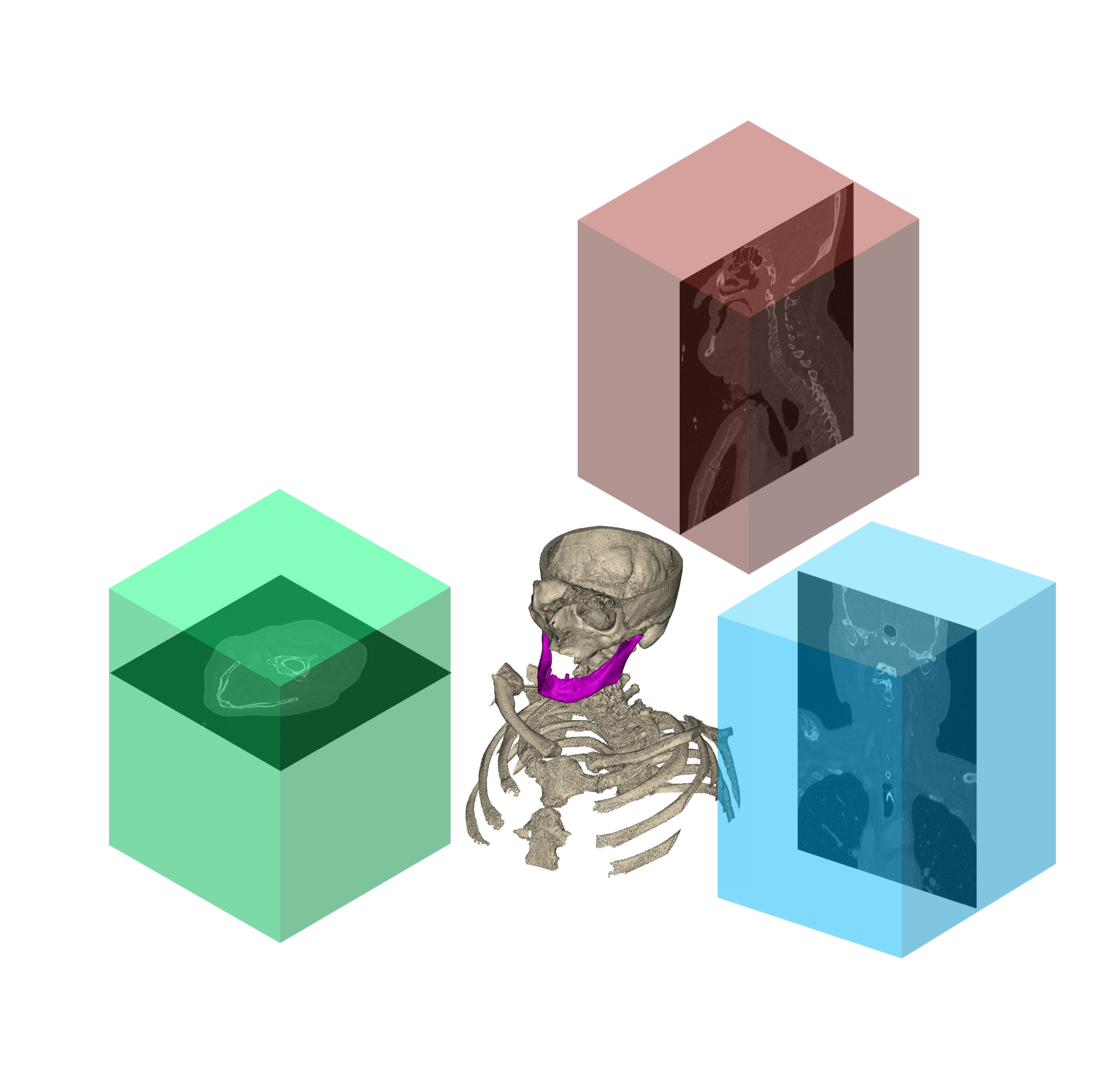}
	\caption{Example of a volumetric CT neck data. Such a scan contains $797$ slices of CT images from the upper chest to the skull of a patient.  }
	\label{fig:data_example}
\end{figure}

In our experiments, we use 11 CT scans reconstructed with a reconstruction kernel of Br64 or I70h(s). Each scan consists of $500$ to $797$ slices with $512\times512$ pixels. We randomly choose eight cases as training, two cases as validation and one case as test. 
%\textbf{ We randomly choose 10 cases data for training and the remaining case for test and further divide the whole training data into $80\%$ training and $20\%$ validation.} 
For ground truth, manual mandible segmentation was performed for each dataset using Materialise Mimics software\footnote{Mimics version 20.0 (Materialise, Leuven, Belgium)} by one trained researcher and confirmed by a clinician. Figure \ref{fig:data_example} shows an example of the data we use for our experiments. 

%{\color{red}{}}

\subsection{Training and test}

The Keras \cite{chollet2015keras} package with the Tensorflow \cite{abadi2016tensorflow} backend is used to train our model on a workstation equipped with Nvidia Tesla K40m GPU of 12GB memory. 
After training the whole network for mandible segmentation, we apply the trained model to the test data. 
The time we used for training one single-channel model 30 epochs is 70 hours in total and the duration for testing the trained model on two various test datasets is 10 minutes. 
%It takes about 70 hours to train one single-direction model for 30 epochs on a workstation equipped with Nvidia Tesla K40m GPU of 12GB memory. 
In the channel network fed with data from the axial plane, we set the input size of the data into $ H=512 $, $ W=512 $ and $ s=3 $. In the Sagittal (Coronal) planes which often have different numbers of slices in every patients, we crop the input images into the same size of $ H= 400$, $ W= 400$ and $ s= 7$ ($ H= 400$, $ W= 400$ and s = 9). We also use Adam optimization \cite{kingma2014adam}  with a learning rate of $ 1 \times 10^{-5} $. The three networks use the same size convolutions with zero padding. The entire architecture and parameters settings of the proposed model are shown in Figure \ref{fig:architecture}. 

\subsection{Experimental results}

\begin{figure}
	\centering
	\includegraphics[width=380pt]{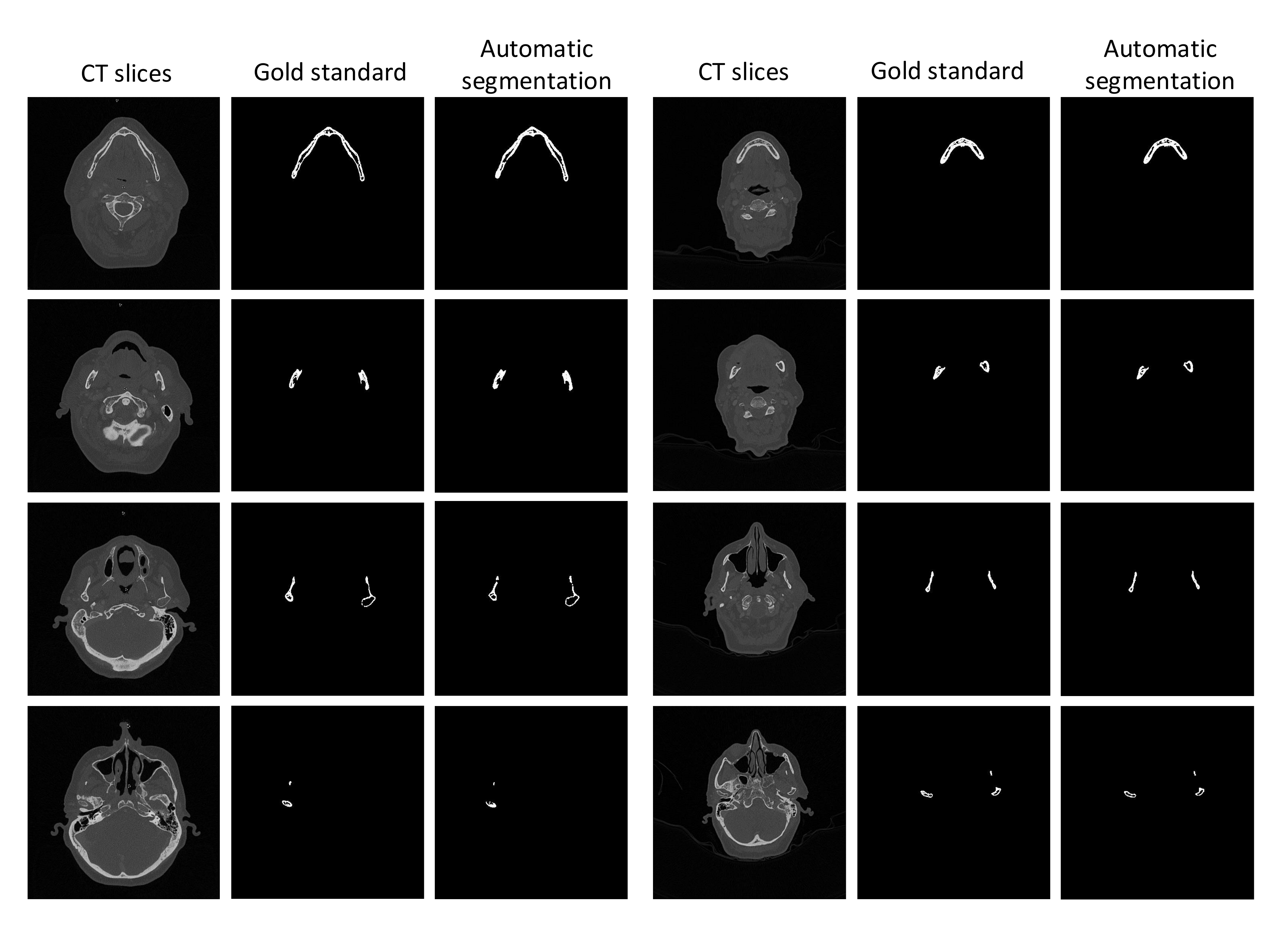}
	\caption{Examples of the automatic segmentation of mandibles in the axial plane. }
	\label{fig:2D_results}
\end{figure}

\begin{figure}
	\centering
	\includegraphics[width=200pt]{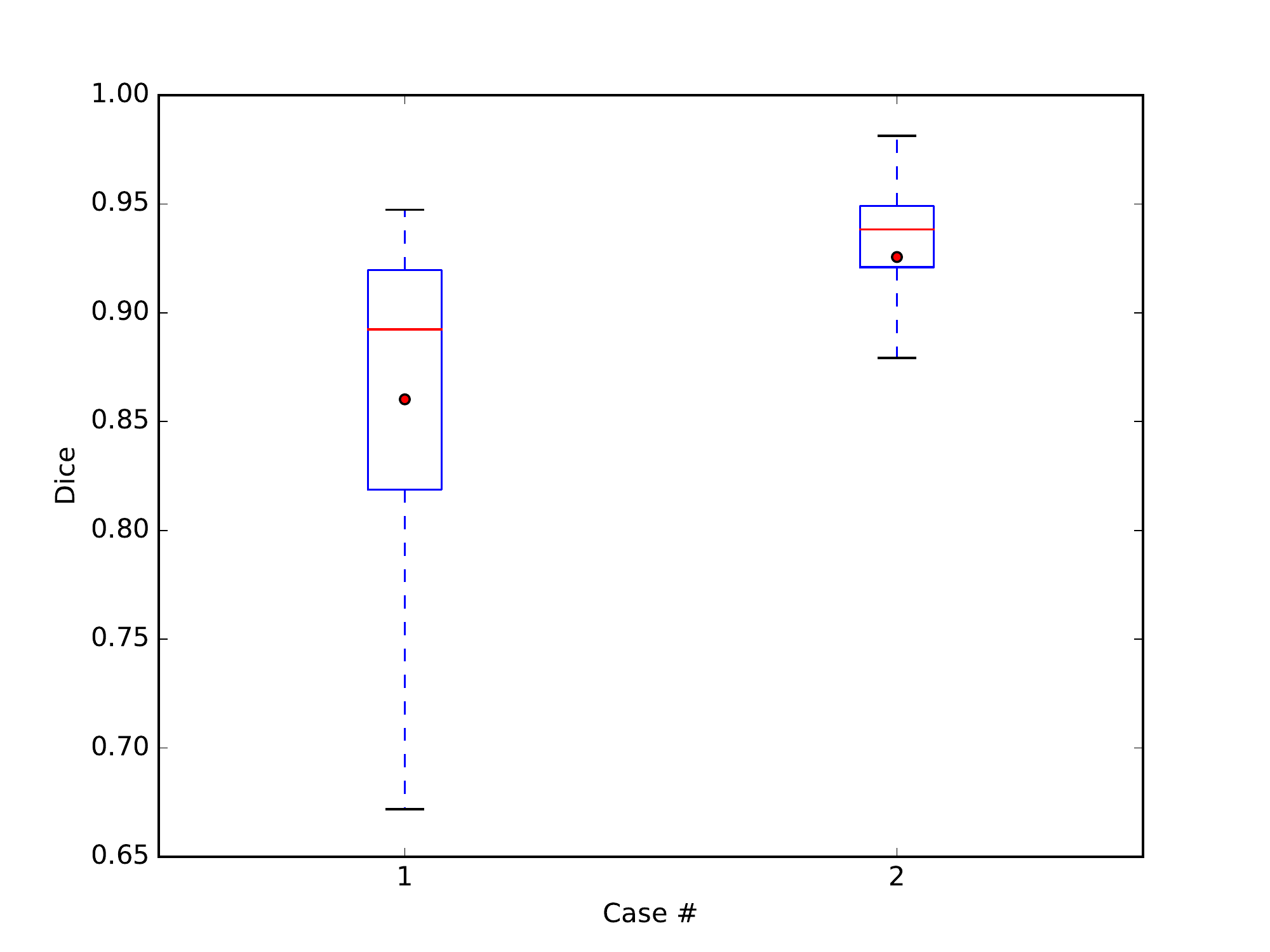}
	\caption{Box-whisker plot of the distribution of the dice coefficience from the test data. The red lines in the middle of the box indicates the median value of the dice while the red points represent the average dice coefficience. }
	\label{fig:tool}
\end{figure}

\begin{figure}
	\centering
	\includegraphics[width=350pt]{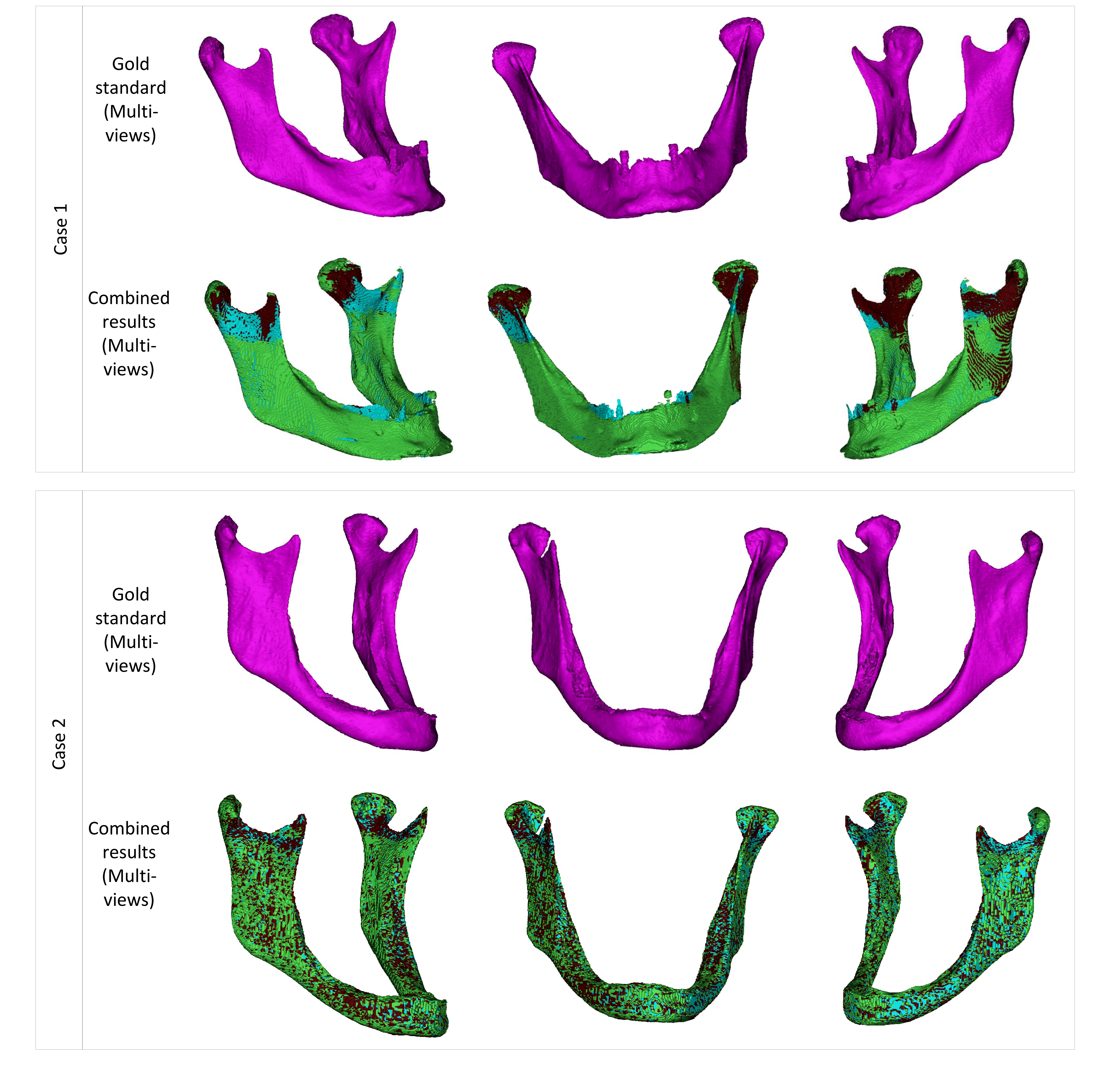}
	\caption{Resulting 3D model of two examples. Every column represents the 3D stacking models in left, middle and right angle of view, respectively. The 3D models in the first(third) row show the golden standard 3D model of the first(second) test case in the different views. The 3D models in the second(fourth) row show the resulting 3D model of the first(second) test case in the different views, in which green, brown and Royal blue mean subnet 1, subnet 2 and subnet 3, respectively. }
	\label{fig:3dmodel}
\end{figure}

%In this subsection, we describe how we evaluate the performance of our proposed approach for mandible segmentation on the 11 scans. 
%We randomly choose 10 cases data for training and the remaining case for test and further divide the whole training data into $80\%$ training and $20\%$ validation and then repeat the experiments two times. 
We repeat the above mentioned process two times and then evaluate the performance of the proposed approach for automatic mandible segmentation. 
%We randomly choose 10 patient data for training and the remaining one case for test and then repeat the experiments two times. 
Figure \ref{fig:2D_results} illustrates some examples of the results achieved by the proposed approach on the test data. 
The average Dice score of two case is $ 0.89 $. The distribution of the Dice score of every slice in two testing examples is illustrated in Figure \ref{fig:tool}. In Figure \ref{fig:tool}, we see that the segmentation accuracy in each image is about $ 0.9 $ comparing to the manual segmentation. This result indicates that the presented method is effective for 3D segmentation of mandible. 
To observe the experimental results visually, we used Materialise Mimics software
%\footnote{Mimics version 20.0 (Materialise, Leuven, Belgium)} 
to stack the 2D segmentation results into a 3D view, as shown in Figure \ref{fig:3dmodel}. 

\section{Discussion}\label{section:discussion}

In this work, we present a CNN-based method for mandible segmentation. 
%There are several improvements contribute to the high performance of our method.
%The performance of our method can be attributed to several factors. 
To the best of our knowledge, this is the first work to demonstrate that mandible in CT images can be effectively segmented using deep learning. In this work, we make full use of the extracted information from different directions to segment the mandible. Our proposed work indicates that 3-dimensional segmentation of medical images can also be achieved by 2D segmentation network which could help avoid the memory issues. Remarkably, this method has a good effect on mandible segmentation, even though until now only a small number of datasets was used during the training. Moreover, instead of using cropped images in some other works for 3D medical image segmentation [15][16], our approach uses the original size of all images, which maximally fits to the computation capacity of GPU and keeps the structural context of mandibles.
%unlike the state-of-the-art 3-dimensional segmentation in the field of medical images \cite{cciccek20163d}\cite{milletari2016Vnet}, the method does not need to rescale the raw data into smaller sizes. 
% so that the network can learn more information from the original CT scans. 
%and we do not need to deal with the computing power issues. 
%, which does not take too much to consider computing power and computer memory. 

%\textbf{Beyond this work, we could extend it from the flowing directions in the future. }
In the future, we could extend this work from the following directions. To begin with, the current implementation treats the corresponding 3D CT images segmentation of a mandible as a video object segmentation with continuous structure, in which we consider the special continuous structure of the mandible. Therefore, the technique of taking into account the neigbouring slices/frames information of each image could be adopted in the application of object segmentation in video frames. 
%Additionally, the next step will be to apply this study method onto MRI data; which is more challenging as the gold standard for bone segmentation is CT and not MRI in the head and neck area.
Moreover, the collection of contextual and shape information from different planes assures the sufficient extraction of useful information from input images, which could provide a future research direction for 3D image segmentation. 
%What's more, collecting features of context and shape in three different directions assures enough useful feature information extracted from the input images, which is also a good research direction for 3D images segmentation. 
Finally, the proposed work could be applied to other organ segmentation or in other imaging modalities, for example, MRI bone segmentation. 

Further evaluation of our approach is required to assess its performance in clinical practice. This could be done through more massive and intensive experiments in or outside our maxillofacial oncology center. Additionally, only 11 patients data are annotated at this moment since the project is still in progress. More experimental datasets would be conducted to verify this technique in the near future. Besides, it is of great importance to verify the automatic mandible segmentation in the 3D virtual planning of craniofacial tumor resection and free flap reconstruction. 
%Although the promising segmentation performance of our method has been evaluated by experimental results in Section \ref{section:experiments}, more evaluations should be carried out to assess its performance in the clinical practice. First of all, it is better if having fully cross-validation to assess the reliability of our method or having enlarging dataset to train and test the model.  Secondly, another effective way to evaluate the resulting segmentation performance is to apply our method to the segmentation of other organizations. Thirdly, it is the most important to verify the bone segmentation performance in the 3D virtual planning of craniofacial tumor resection and free flap reconstruction. 

\section{Conclusion}\label{section:conclusion}

This paper proposed an end-to-end approach for automatic segmentation of mandible in CT scans. Such an approach has three channels for 2D segmentation of mandibles in multisectional CT scans and then combines the resulting 2D segmentation into 3D segmentation. We implement the proposed approach on 11 neck CT scans and achieve an average dice coefficient of $0.89$. The experimental results demonstrate the effectiveness of the proposed approach in mandible segmentation and its potential employment in 3D virtual planning of craniofacial tumor resection and free flap reconstruction. 

%This paper proposed a novel technique based on deep learning for segmentation of mandible. The proposed method learned an end-to-end mapping between the original CT images and the annotated label images through training 10 cases and testing 1 case. With the proposed neural network framework, a 3D segmentation model can be obtained directly by extracting feature information from its corresponding CT scans without down-sampling. The experimental results demonstrate that our method can segment mandible in original CT effectively and produce high-quality results. 
%\subsubsection*{Acknowledgments}
%
%Use unnumbered third level headings for the acknowledgments. All
%acknowledgments go at the end of the paper. Do not include
%acknowledgments in the anonymized submission, only in the final paper.
%\newpage
\section*{References}
\bibliographystyle{dcu}
\bibliography{MIDL_using_nips_2017_for_arxiv}

\end{document}